\documentclass[sigconf, review = false]{acmart}

\usepackage{listings}
\usepackage{multicol}
\usepackage{algorithm}
\usepackage{algpseudocode}
\usepackage{listings}
\usepackage{multirow}
\usepackage{comment}
\usepackage{amsthm}
\usepackage{amsmath,amsfonts}
\usepackage{tabularx} 

\usepackage{graphicx}
\usepackage{subcaption}
\usepackage{geometry}

\usepackage{academicons}
\usepackage[page]{appendix} 

\makeatletter
\newcommand{\SplitState}[1]{%
  \State
  \parbox[t]{\dimexpr\linewidth-\ALG@thistlm}{%
    #1\par\xdef\Split@prevdepth{\the\prevdepth}%
  }\par
  \prevdepth\Split@prevdepth
}
\makeatother
\begin{document}




\title{Enhancing Model Fairness and Accuracy with Similarity Networks: A Methodological Approach}


\author{Samira Maghool}

\author{Paolo Ceravolo}
\begin{abstract}
In this paper, we propose an innovative approach to thoroughly \textit{explore dataset features} that introduce \textit{bias} in downstream \textit{machine learning} tasks. Depending on the data format, we use different techniques to map instances into a similarity feature space. Our method's ability to adjust the resolution of pairwise similarity provides clear insights into the relationship between the dataset classification complexity and model fairness. Experimental results confirm the promising applicability of the similarity network in promoting fair models. Moreover, leveraging our methodology not only seems promising in providing a fair downstream task such as \textit{classification}, it also performs well in \textit{imputation} and \textit{augmentation} of the dataset satisfying the fairness criteria such as demographic parity and imbalanced classes.
\end{abstract}


\maketitle

\section{Introduction}

Machine learning (ML) algorithms rely on a substantial amount of training data, assuming that it accurately represents reality and is free from biases introduced during data collection. However, every dataset inherently carries biases from its social context of origin and may suffer from fundamental quality issues~\cite{naeem2021exploration}, leading algorithms to produce results that favor one group over another. When these group distinctions are based on sensitive features, the model ends up reflecting social biases rather than learning the true underlying rules, resulting in discriminatory decisions.
A common approach to address this issue is to modify the data distribution or representation within the dataset \cite{mehrabi2021survey}, such as balancing groups \cite{fish2016confidence} or removing sensitive features \cite{agarwal2023fairness}. However, it should be noted that these steps can potentially impact the prediction accuracy of algorithms and their ability to unveil hidden relationships, patterns, and latent associations in the data. Therefore, the literature looks to find a delicate balance between fairness and accuracy by considering the specific context, domain expertise, and potential consequences of data alterations \cite{fish2016confidence,liang2022algorithmic,zhang2021balancing}. Researchers have developed techniques like fairness-aware learning algorithms, which integrate fairness metrics during model training to strike a better balance between fairness and accuracy \cite{mary2019fairness, krasanakis2018adaptive}. Application of the above-mentioned methods needs to take into consideration intersectionality; which is mostly about considering multiple sensitive attributes that might intersect (e.g., race and gender) and is not feasible to consider using the common methods.

In previous works[omitted due to anonymity], we introduced a novel solution, for debiasing the datasets aimed at achieving fair and accurate ML models. Our methodology involves a sequence of mappings on the dataset transforming it into a similarity feature space using different tuning techniques such as \textit{Scaled exponential kernel} \cite{smola2003kernels} and \textit{Random Walk kernel} \cite{NIPS2015_31b3b31a}. 
This secondary dataset is then represented as a \textit{Similarity Network}, which is interpretable due to its graph structure and usable for downstream ML tasks. In this paper, we provide a comprehensive overview of our methodology, detailing the steps to be followed and the range of methods that can be employed. This way we highlight the relationship between these methods and their respective applications, showing how the choice of method can be tailored to the specific characteristics of the dataset under study and the intended downstream tasks.

This paper is organized as follows:
Section \ref{sec:m&m} presents our methodology implementation and later provides a list of applications for validating the secondary dataset (in Section \ref{app}). Section \ref{eval} covers analytical tasks and evaluation metrics. Section \ref{sec:exp} explores the effectiveness of our methodology for one of the listed applications. Finally, Section \ref{sec:conc} concludes, highlighting achievements and outlining future research challenges.

\section{The Proposed Methodology }\label{sec:m&m}

\begin{figure}[ht]
  \centering
  \includegraphics[width=\linewidth]{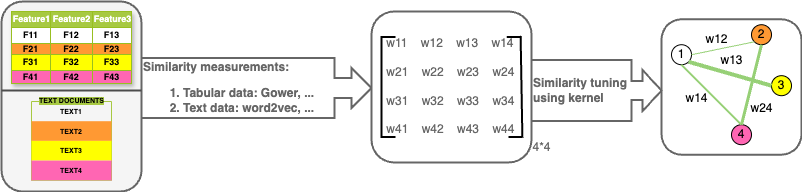}
  \caption{ (I) Graph construction from datasets based on instances' similarities. (II) An $N\times N$ adjacency matrix demonstrates the links' weights in the similarity network. (III) The final graph represents dataset entries as a weighted network $\mathcal{N} = \textbf{(V, E)}$, where $\mathbf{V}$ corresponds to the network nodes (vertices), and $\mathbf{E}$ to the links (edges) among them.\label{fig:flow}}
\end{figure}

In order to implement our methodology, we proceed according to the three steps illustrated in Figure \ref{fig:flow}. 
In continuation, we explain in detail how to build the Similarity Network and tune the weight of links using kernel functions inspired by the method explained in \cite{wang2014similarity}.

\subsection{Building the Similarity Network and tuning by kernels.}

According to the type of data set, we choose the proper method for measuring the pairwise similarity between instances. For the tabular data, containing both categorical and numerical values, \textit{Gower Distance  $\mathbf{(GD)}$} method  ~\cite{gower1971general} or other distance measurement techniques could be used to calculate the distances so that the similarity is acquired as $1-\mathbf{GD}$. Once the dataset contains text in documents as dataset instances, Natural Language Processing (NLP) is useful for constructing the similarity network of data points. Using algorithms such as \textit{word2vec} \cite{mikolov2013distributed}, nodes are
embedded in a latent space, considering the similarity of their features in
the real space.

Given all the pairwise similarities between the dataset entries, defining a similarity network requires selecting a threshold value that determines the presence or absence of links between instances. 
Choosing an inappropriate threshold can introduce unfavorable biases to the model. 
To tackle this challenge, kernel matrices that capture the local and global topological characteristics of the underlying graph may offer a solution.
Utilizing \textit{scaled Exponential kernel (Ek)}, and \textit{Random Walk kernel (RW)} (see more details in the Appendices \ref{sim_measurements} and \ref{sim_tune}) provides the possibility to tune the weight of features for creating the similarity network while reaching an effective balance between accuracy and fairness. Algorithm \ref{app.1} in Appendix \ref{app.2} also demonstrates the required steps to proceed in creating the similarity network to be used for further applications as described in Section \ref{app}.

\subsection{Applications of the proposed algorithm}\label{app}
\begin{figure}[h]
    \centering
    \includegraphics[width=0.35\textwidth]{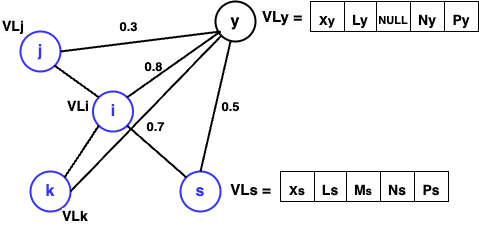}
    \caption{\small{Imputing the missing value (NULL) process using the features of similar nodes.}}
    \label{fig:imp}
\end{figure}

In this section, we briefly point out some applications of the transformed dataset resulting from our methodology:\\
\textbf{Classification:} Classification problems may encounter different levels of complexity based on the dataset characteristics (see Section. \ref{ccm} for more details on these metrics). Some of these characteristics such as \textit{Feature-based}, \textit{Linearity}, and \textit{Class imbalance}  measures could be calculated using \cite{lorena2019complex}, represent how the classification of a given dataset based on its features could be challenging. The presence of sensitive features, among the most important ones, could signal bias in the classification output once the dataset is easily classified based on a few features. 
As it will be evaluated in Section \ref{sec:exp}, the comparison between the output of classification on the original dataset with the mapped dataset using our methodology, demonstrates improvements in fairness and accuracy metrics despite the higher level of complexity of the dataset.\\
\textbf{Data imputation:} Considering the deficiencies of the existing datasets, mostly related to missing data points during the data collection process, we have to choose the appropriate methodology to impute those missing values for training ML models. The algorithms in use for data imputation purposes mostly disregard the sensitivity of the missed feature values, and fill them with overall \textit{mean} or \textit{the most frequent} values which in some case scenarios this choice will lead to amplifying the existing bias in the dataset. Once the original dataset is debiased using our proposed method, a missed value could be imputed using features of the nodes closely located in the similarity space. Figure \ref{fig:imp} demonstrates schematically the dataset entries as nodes that are linked together through weighted edges. Each node's features are demonstrated as a Vector-Label (VL). Using our approach, the \textit{NULL} values could be imputed using the values of the relevant features in the vicinity using an aggregation function $\mathbf{F(\{M_{l}\})}, \mathbf{l} \in [neighbor\mathbf{(y)}: i, j, s, k]$ to calculate the $\mathbf{M_{y}}$ in the Figure \ref{fig:imp}. In a simple scenario, we can consider the $\mathbf{F}$ as a \textit{mean} for numerical values of feature $\mathbf{M}$ or the \textit{majority} once the values are categorical.   \\

\noindent
\textbf{Data augmentation:} In addressing the limited availability of data for predictive purposes with ML, there are concerns about potential biases arising from dataset augmentation using traditional methods. Despite advanced algorithms to generate synthetic data that preserve the original data distribution, challenges remain, including the risk of perpetuating social biases. Our approach can strategically generate synthetic data points relatively close to nodes in the structured similarity network. A \textit{Vector-Label Propagation} algorithm such as [\texttt{doubleblind}], complemented by a kernel function for adjusting links' weight, is utilized to label these synthetically generated points. The primary goal is to reduce the system’s dependence on sensitive features for ML tasks without excluding them, thereby avoiding the risk of exacerbating biases or reducing data variation. Implemented in a big data ecosystem, our methodology enables continuous evaluation in an evolving domain, effectively addressing the challenges of data scarcity with a fairness-aware approach. In [\texttt{doubleblind}] we have extensively discussed our methodology for data augmentation. It has been able to debiase the dataset and improved the fairness metrics of a classification model.
Considering the diverse applications of our proposed methodology, in Algorithm \ref{app.3}, we have presented a pseudo-code to make it more clear. 
\begin{figure}[h]
    \centering
    \includegraphics[width=0.48\textwidth]{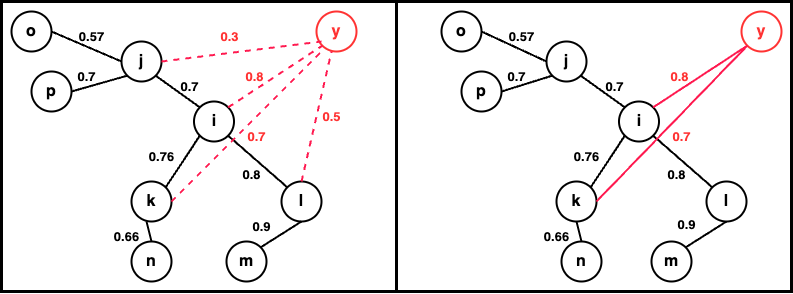}
    \caption{\small{Schematic view of data augmentation proposed as an application of created similarity network. After deciding about the threshold in keeping the links, the newly added node, y, is linked to nodes i and k with the largest link's weight. The $VL_{y}$ will be resulted from a Vector-Label Propagation} algorithm while having eyes in the fairness metrics to be evaluated.}
    \label{fig:aug}
\end{figure}

\section{Evaluation metrics}\label{eval}

To assess the equilibrium between accuracy, fairness, and complexity, we employ a range of evaluation metrics in our analysis.

\begin{enumerate}
    \item \textbf{Accuracy:} Among the standard metrics evaluating the predictive algorithms, we can mention, the F1 score, Accuracy, and Recall~\cite{Japkowicz2015}. 
    \item \textbf{Transparency:}\label{trans} Regarding the model output transparency and explainability, we adopted the SHAP algorithm~\cite{lundberg2017unified}. Using this algorithm we can interpret how a model predicts a label based on the data set features.
    \item \textbf{Classification complexity measures:}\label{ccm} Considering the fact that classification performance metrics do not provide a full insight into the level of complexity of a classification problem that an algorithm has to deal with, we need to investigate whether the failure to distinguish different classes of the data set is caused by the incapability of a model in taking into account the crucial factors, or the data set intrinsically does not allow a sharp separation between classes. In simple words, we need to see how a classification problem is complex in nature and if some of the features play a more effective role than others. 
A classification model may bring up a highly accurate model by considering sensitive features during classification. In this regard, we highlight that, if a class is distinctively separated by a subset of sensitive features, this could be a sign of bias or discriminative behavior imposed on the system (data set or model or both).
%
%
%
We examine these measurements for different representations of the dataset in the similarity feature space using different kernels to figure out how the kernels could change the complexity and accuracy of a model. 

According to ~\cite{lorena2019complex}, classification complexity measures could be categorized as follows:

\textbf{Feature-based measures} characterizes the discriminative power of the available features; F1, F1v, F2, F3, F4 belong to this category.
    
\textbf{Linearity measures} try to quantify whether the classes can be linearly separated. From this category we can name L1, L2, L3.
    
\textbf{Neighborhood measures} which characterize the presence and density of the same or different classes in local neighborhoods; N1, N2, N3, N4, T1, LSC are defined for this purpose.

\textbf{Network-based measures} extract structural information from the data set by modeling it as a graph and measure the Average density of the network (Density), Clustering coefficient (ClsCoef), Hub score (Hubs).
    
\textbf{Dimensionality measures}, which evaluate data sparsity based on the number of samples relative to the data dimensionality; measures in this category are, for example, T2, T3, T4.
    
\textbf{Class imbalance measures} consider the ratio of the number of examples between classes. C1 and C2 quantify the class imbalances.
\item \textbf{Fairness:}\label{fairness} To examine the presence of bias in the system (both in the dataset and the model), we follow a three-step approach.
(i) We consider complexity measures to assess whether the dataset is heavily characterized by specific features, which might manifest as high values in the feature under analysis.
(ii) By analyzing the most significant features identified by a classifier, we monitor whether these features correspond to sensitive attributes.
(iii) We employ SHAP (SHapley Additive exPlanations) values to gauge feature importance and to understand how distinctively each feature influences the model.

In the final stage, to quantify whether these features introduce bias into the system, we distinguish between \textit{Privileged} and \textit{Unprivileged} classes, based on specific feature values related to sensitive attributes. The privileged class comprises instances with protected or sensitive features, such as race or gender, that may play a role in achieving favorable outcomes. For instance, a privileged class could be defined by attributes like {\tt Race}: \textit{white}, {\tt Marital status}: \textit{married}, and {\tt Gender}: \textit{male}.

In evaluating fairness, we explore whether privileged and unprivileged groups have \textbf{Equal Opportunity} (Eq. \ref{eq}) and \textbf{Equal Mis-opportunity} (Eq. \ref{miseq}) in the model's predictions~\cite{oneto2020fairness}. This can be computed by:

\begin{equation}\label{eq}
    Pr(Y^{\prime}=1\mid s=1,Y=1) - Pr(Y^{\prime}=1\mid s=0,Y=1),
\end{equation}

\begin{equation}\label{miseq}
    Pr(Y^{\prime}=1\mid s=1,Y=0) - Pr(Y^{\prime}=1\mid s=0,Y=0),
\end{equation}

where $Y$ represents the ground-truth label, $Y^{\prime}$ denotes the predicted label, and $s$ is a binary indicator distinguishing privileged ($s=1$) and unprivileged ($s=0$) groups. Consequently, \textit{Equal Opportunity} assesses the equality of True Positive (TP) rates, effectively measuring the equality of recall values for privileged and unprivileged groups. In contrast, \textit{Equal Mis-opportunity} focuses on accuracy, specifically the False Positive (FP) rate, and aims to ensure it is equivalent for privileged and unprivileged classes.
In practice, these measures evaluate whether the classifier's outputs for privileged and unprivileged classes tend to assign fair labels to each class. Therefore, the closer the results of Eq. \ref{eq} and Eq. \ref{miseq} are to zero, the fairer the model's treatment of sensitive features.
\end{enumerate}

\section{Experimental Results}\label{sec:exp}


In order to demonstrate the effectiveness of the proposed methodology, in our recent paper[\texttt{doubleblind}], we used a public data set
of the Curriculum Vitae (CV) of 301 employees\footnote{https://rpubs.com/rhuebner/hrd cb v14
} which contains both numerical and categorical features such as {\tt Zipcode}, {\tt Age}, {\tt Gender}, {\tt Marital
Description}, {\tt Citizen Description}, {\tt Hispanic/Latino}, {\tt Race}, {\tt Employment Status}, {\tt Department},
{\tt Position}, {\tt Payment}, {\tt Manager Name}, {\tt Performance Score},
and, {\tt Work experience}.
Our primary objective is to investigate the capability of similarity networks to enhance fairness in ML tasks. To achieve this, we use Gower distance ($\mathcal{GD}$), and for defining the similarity we calculate $\mathcal{SGD} = 1-\mathcal{GD}$. To capture latent patterns and relationships in the dataset, we tune the quantified similarity using kernels. To evaluate the ability of our approach to leverage the network representation and the impact of the kernel matrix in capturing the main characteristics of the dataset while keeping the explainability of the output effectively, we considered the following points:

I) We implemented a 10-fold cross-validated classification model using Random Forest (RF) and XGBOOST to predict the \textit{Pay Rate} categories on the original dataset and different representations of it. Since this is a multi-class classification problem, we consider the F1 weighted score to avoid bias in evaluating the output in favor of the highly populated class (\ref{tab:1}). 
\begin{itemize}

\item The first row displays the performance of the classifiers on the preprocessed original dataset. 
\item In the second row, we implemented the classification on the class-balanced dataset using SMOTE\footnote{https://imbalanced-learn.org/stable/references/generated/imblearn.over\_sampling.SMOTE.html} technique. Using this technique, we over-sampled the instances in Pay Rate classes in the minority. 
\item The third and fourth rows show the results for balanced groups, differentiated by protected characteristics such as gender and ethnicity. Basically, we oversampled the smaller group based on gender/ethnicity by adding minimal variations to the numerical values while leaving the categorical values as they were.
\item Additionally, we evaluate the performance of the classifiers on the similarity values of the instances using $\mathcal{SGD}$. For each dataset entry ($a$), we considered a set of normalized pairwise similarity values (as computed using Eq. \ref{form2}) to other entries ($b$) as the feature values in the mapped space.

\item Finally, we apply both classifiers to the mapped similarity graph using kernels, Ek (Eq. \ref{form3}) and RWk (Eq. \ref{rwk}). 
    
\end{itemize}
II) To measure the complexity level of the classifier domains; feature-based, linearity, neighborhood, network, dimensionality, and class imbalance, we used the
methodology explained in \cite{lorena2019complex}. Where the complexity measures are defined for binary classification problems, we have used the One-Vs-One multiclass strategy (OVO) to study the classification complexity of the dataset. For most of the criteria, we observed that the classification complexity of the original dataset is low, showing that the data can be easily classified based on a few features. Then, as discussed in detail in our previous paper [\texttt{doubleblind}], we considered three data representations using different mapping strategies: $\mathcal{SGD}$, $\mathcal{SGD}+$Ek, $\mathcal{SGD}+$RWk, to explore how the classification complexity measures change. We observed that although the performance of the classifiers is greatly increased by mapping the original dataset into the similar feature space, the complexity measures still decrease compared to the original dataset, and this reduction becomes even more obvious after using $\mathcal{SGD}+$Ek. Implementing the $\mathcal{SGD}+$RWk mapping not only keeps the performance acceptably high, but also increases the level of complexity of the classification in the mapped dataset, which means that classification by just a few features is no longer easily possible.

\begin{figure}[h]
    \centering
    \includegraphics[width=0.5\textwidth]{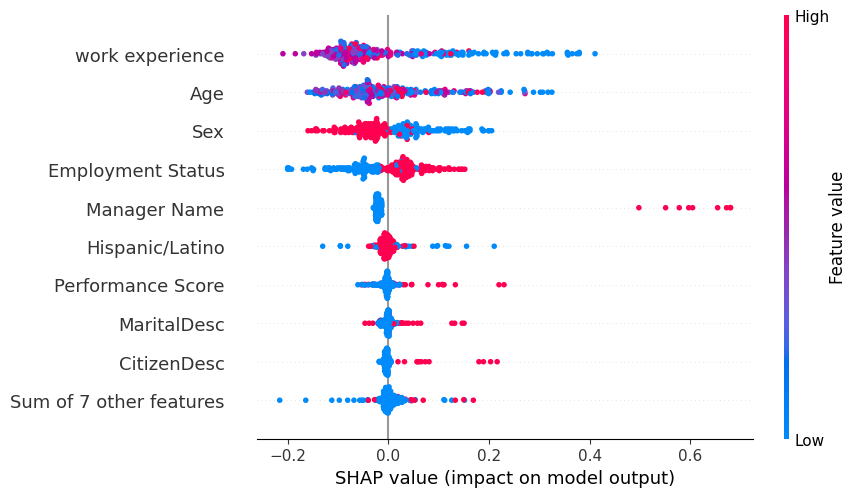}
    \caption{\small{The most important features SHAP values, extracted from a RF classifier predicting the pay rate.}}
    \label{fig:shap}
\end{figure}

III) In Fig. \ref{fig:shap} we show the most important features and their corresponding SHAP values. Red/blue dots indicate the high/low value of a particular feature that affects the output of the model. We see that sensitive features such as {\tt Gender}, {\tt Age}, and {\tt Hispanic/Latino} not only have a large impact on the output of the model, but at some point the imposed discrimination is obvious. We then decided to use them to create privileged/unprivileged groups.

IV) To calculate fairness metrics, we need to consider binary classification. For this purpose, we categorized the \textit{Pay Rate} into two levels. According to Fig. \ref{fig:shap}, we considered {\tt Gender} and {\tt Age} as two discriminating features, defining the privileged class as {\tt Gender}: \textit{male} and {\tt Age} $\leq$ \textit{40}, while considering the remaining instances as the unprivileged class. We measure the true positive rate (TPR) and the false positive rate (FPR) in a basic prediction model. The closer the values are to zero, the less biased the system is. In this experiment, measuring Equal Opportunity/Mis-Opportunity, we get $0.3/0.17$ for the original data set, while these values are reduced to $0.25/0.027$ and $0.2/0.02$ for $\mathcal{SGD}$ and $\mathcal{SGD}+$Ek, respectively. 

V)
In [\texttt{doubleblind}], we extend the capabilities of our methodology by using the density of the network to identify less represented groups within the dataset for data augmentation purposes. Generating synthetic data near these groups we naturally balance the representation of different groups in the dataset. 

To ensure continuous verification, we deployed the training and verification pipelines of the ML model using a platform in 
[\texttt{doubleblind}].
This platform provides an edge-cloud continuum service infrastructure that delivers high-performance computing resources while assuring advanced non-functional properties such as privacy and application security. The solution proposed by [\texttt{doubleblind}] utilizes contract-based continuous verification using evidence gathered through transparent monitoring. In this way, we have evaluated whether the dataset augmentation method we propose is able to enlarge the data points while guaranteeing fairness in the ML models trained after the augmentation.

Implementing the augmentation procedure, the $C2$ measure computed for measuring class balance demonstrates an improvement by $60\%$ in balancing the classes using our method. The Collective Feature Efficiency ($F4$) measure to get an overview of how the features work together, lower values of $F4$  indicate that it is possible to discriminate more examples and, therefore, that the problem is simpler. Our measurements show $F4$ increased by $34\%$ which means the classification problem is more difficult in augmented data due to the debiasing process as we expected. To explore the applicability of our method to improve fairness by creating synthetic data points, we have measured Equal Opportunity and Equal mis-opportunity. The measured metrics on the original data and after augmentation demonstrate improvements from $0.35/0.17$ in the original dataset to $0.17/0.08$ in the augmented version by synthetic data entries.

\begin{table}[]
\centering
\resizebox{!}{2cm}
{
\begin{tabular}{|c|l|l|}
\hline
\multicolumn{1}{|l|}{\textbf{\begin{tabular}[c]{@{}c@{}}Dataset representations\end{tabular}}}& \textbf{Random Forest}       & \textbf{XGBOOST}             \\ \hline
\multicolumn{1}{|l|}{\textbf{\begin{tabular}[c]{@{}c@{}}Original data set\end{tabular}}}                               & $0.666 \pm 0.056$ & $0.680 \pm 0.059$
\\ \hline
\multicolumn{1}{|l|}{\textbf{\begin{tabular}[c]{@{}c@{}}Original data set\\ with balanced classes\end{tabular}}}                               & $0.842 \pm 0.038$ & $0.817 \pm 0.046$

\\ \hline
\multicolumn{1}{|l|}{\textbf{\begin{tabular}[c]{@{}c@{}}Original data set with balanced\\ protected feature (gender)\end{tabular}}}                               & $0.700 \pm 0.071$ & $0.686 \pm 0.067$
\\ \hline
\multicolumn{1}{|l|}{\textbf{\begin{tabular}[c]{@{}c@{}}Original data set with balanced\\ protected feature (ethnicity)\end{tabular}}}                               & $0.783 \pm 0.049$ & $0.777 \pm 0.055$
\\\hline
$\mathcal{SGD}$                                                                   & $0.944 \pm 0.039$                & $0.938 \pm 0.034$                \\ \hline
\textbf{\begin{tabular}[c]{@{}c@{}}Exponential kernel\\  of $\mathcal{SGD}$\end{tabular}}  & $0.921 \pm 0.039$                & $0.926 \pm 0.048$                \\ \hline
\textbf{\begin{tabular}[c]{@{}c@{}}Random Walk kernel\\ of $\mathcal{SGD}$\end{tabular}}   & $0.888 \pm 0.055$                & $0.907 \pm 0.051$           
         \\ \hline
\end{tabular}}\caption{\small{The mean $\pm$ standard deviation of F1 weighted score using XGBOOST and RF classification algorithms on the CV data set considering original and class-balanced data set; on balanced group data set respectively gender and ethnicity; and different representations of the dataset; similarity network using $\mathcal{SGD}$, mapped data by the Ek and, RW kernels.}}\label{tab:1}
\end{table}

\section{Conclusion}\label{sec:conc}
In this paper, we aim to propose a comprehensive and innovative approach to explore and address the possible bias of the dataset against groups of instances. For this purpose, local and global similarities between instances are measured using different algorithms and tuned to maintain the delicate trade-off between accuracy and fairness of an ML model. In this work, we clarify the steps to be followed and the variety of methods that can be exploited based on specific properties of the studied dataset and the downstream applications of the methodology. By addressing the challenges associated with data scarcity, our method contributes to the ongoing quest to develop unbiased ML systems that can effectively generalize across diverse and dynamic datasets. While our work has provided valuable insights, it is important to acknowledge a primary limitation.
In this work, we conducted our experiments exclusively on a single public dataset, which means that our results may be somewhat dependent on the specific problem under investigation. Future research will focus on corroborating our conclusions by experimenting with different public datasets.



\bibliographystyle{ACM-Reference-Format}
\bibliography{main.bib}

\newpage
\appendix
\begin{appendices}

\section{Basic Notions}
\subsection{Similarity measurements}\label{sim_measurements}

    Graph construction from datasets based on instances' similarities can be implemented using various mechanisms to measure similarity values. In this paper, we consider two methods for calculating data point similarity and quantifying it to build the weighted network $\mathcal{N} = \textbf{(V, E)}$, where $\mathbf{V}$ corresponds to the network nodes (vertices), and $\mathbf{E}$ to the links (edges) among them.

\begin{enumerate}
    \item \textbf{Average Normalized Similarity:} According to~\cite{abdel1984monte,gliozzo2022heterogeneous}, when data is represented by features with homogeneous numeric types, pairwise similarity is generally quantified by Pearson correlation, cosine similarity, or the inverse of the Euclidean distance. Otherwise, when the data points are represented by features with heterogeneous types, the \textit{Gower distance} ($\mathcal{GD}$) is often used~\cite{gower1971general}. More precisely, if $\mathbf{a}$ and $\mathbf{b}$ are the values of the variable $g$ for data points $\mathbf{a}$ and $\mathbf{b}$, and if $\mathbf{G}$ is the set of all values for the variable $g$ (e.g., age), the (potentially normalized) similarity between points $\mathbf{a}$ and $\mathbf{b}$ according to feature $g$ is computed as:
    \begin{equation}\label{sim}
        S(a, b, g) = 1-\frac{abs(a-b)}{max(G)-min(G)}
    \end{equation}
    if $g$ has a numeric type, or by:
    \begin{equation}\label{sim}
        S(a, b, g) = I(a==b)
    \end{equation}
    if $g$ has a categorical type, where $I(x)$ is the indicator function that returns 1 if the logical expression $x$ is true (other similarity measures may be used depending on the problem at hand). For a set of $k$ variables $\mathbf{G}$=\{$g_{1},g_{2},..g_{k}$\}, the Gower similarity $\mathbf{S^\prime}$ between two entries $\mathbf{a}$ and $\mathbf{b}$ is then defined as the average of the (normalized) similarities for each of the variables:
    \begin{equation}\label{form2}
        S_{a,b}^\prime = \frac{\sum^{k}_{i=1}S(a, b, g_{i})}{k}
    \end{equation}

    \item \textbf{Using NLP in Similarity Measurement:} As a second approach for measuring similarity, we leverage Natural Language Processing (NLP) to construct the similarity graph of data points. Using this technique, agents are embedded in a latent space, considering the similarity of their features in the real space.

    In this approach, feature vectors are created for each node as a tokenized document and then transformed by an embedding algorithm into a low-dimensional space. The \textit{word2vec}~\cite{mikolov2013distributed} algorithm employs a shallow neural network to learn word-associated features from a large corpus of text, resulting in a set of low-dimensional vectors assigned to each word. In this way, similar nodes are mapped closer in the latent space.
    Using the Gensim\footnote{https://radimrehurek.com/gensim/models/word2vec.html} package for node embeddings, the similarity of two node instances (i.e., nodes representing data points $\mathbf{a}$ and $\mathbf{b}$), featuring their characteristics in the original dataset, is computed using cosine similarity.
\end{enumerate}

\subsection{Similarity tuning by kernels}\label{sim_tune}



Given all the pairwise similarities between points in our dataset, the definition of a similarity network requires choosing the value of the $\epsilon$ parameter that defines the existence (non-existence) of links between points. 
Selecting the proper threshold value is a crucial task that surely affects the algorithm's performance. Moreover, this choice may also introduce unfavorable biases to the model; indeed, too small values could easily result in too sparse networks, i.e. networks composed of many connected components (subgraphs), while too large values would increase the network connectivity, resulting in spread information and impacting on the computational cost required to leverage this graph data for further analysis.
To address this issue, kernel matrices that are able to capture the topological characteristics of the underlying graph may provide a solution. 
Edge weights are represented by an $n \times n$ similarity matrix $\mathbf{W}$ where $\mathbf{W(a, b)}$ indicates the similarity between data points $\mathbf{a}$ and $\mathbf{b}$. Hereafter, we denote $\mathbf{\rho(a, b)= 1-S^\prime(a,b)}$ as the distance between data points, as computed by one of the similarities described in Sec. \ref{sim_measurements}. Initially, we use a scaled Exponential kernel (Ek) to determine the weight of the edges:
\begin{equation}\label{form3}
   W(a,b) = exp(-\frac{\rho^{2}(a, b)}{\mu \epsilon_{a,b}}),
\end{equation}

where according to the ~\cite{wang2014similarity}, $\mu$ is a hyperparameter that can be empirically set and $\epsilon_{a, b}$ is used to eliminate the scaling problem by defining:

\begin{equation}\label{eq:eps_SNF}
    \epsilon_{a,b} = \frac{mean(\rho(a, N_{a}))+mean(\rho(b, N_{b})) + \rho(a,b)}{3},
\end{equation}

where $mean(\rho(a, N_{a}))$ is the average value of the distances between $a$ and each of its $k$ NNs. The range of $\mu = [0.3, 0.8]$ is recommended by ~\cite{wang2014similarity}.

In the next step, the kernel matrix $\mathbf{K}$ is derived from the similarity matrix $\mathbf{W}$. To this aim, we used the Random Walk kernel (RWk) which is calculated as:

\begin{equation}\label{rwk}
    K=(m-1)I + D^{-\frac{1}{2}}WD^{-\frac{1}{2}},
\end{equation}

where $\mathbf{I}$ is the identity matrix, $\mathbf{D}$ is the diagonal matrix with elements $d_{ii}=\sum_{j} W_{ij}$ and $m$ is a value greater than $2$. A p-step random walk kernel can be achieved by multiplying $\mathbf{K}$ by itself for the $p$ times. RWk has the effect of strengthening (diminishing) similarities that were already high (low) in the input matrix $\mathbf{W}$, therefore highlighting only the relevant information.

\section{Methodology implementation}\label{app.2}

In this section, we provide the pseudo-codes for implementing the proposed methodology. In the Algorithm \ref{app.1} we define the steps to be followed for creating the Similarity Network while in the Algorithm \ref{app.3}, we have presented the task-wise procedure for using the mapped dataset in the similarity space for upcoming fairness-based tasks.
\begin{algorithm*}[ht]
\caption{Algorithmic view of the methodology regarding the upcoming applications; Calculating the similarity}\label{app.1}
\begin{algorithmic}
\Require input dataset (D); define similarity measurements (S); kernel functions (EK, RWK) hyper-parameters; performance and, fairness metrics (DF, TPR, FPR); fairness threshold ($\delta$), list of applications A= [\textit{Classification (Clf), Imputation(Imp), Augmentation (Aug)}]
\State \textbf{1) Calculate the similarity.}
\State N: Number of instances in the dataset.
\If{D contains categorical/numerical values}
\State i) Calculate pairwise Gower Similarity $\mathbf{S^\prime}$,
\SplitState{ii) Using EK, RWK for similarity tuning to be used as\\ network's edges ($w_{ij}$) between nodes $i$ and $j$. }
\ElsIf{D contains textual values}
\State i) Measure the $\mathbf{S^\prime}$ using NLP algorithms, 
\SplitState{ii) Using EK, RWK for similarity tuning to be used as \\
network's edges ($w_{ij}$) between nodes $i$ and $j$. }
\EndIf

\end{algorithmic}
\end{algorithm*}

\begin{algorithm*}[t]
\caption{Algorithmic view of the methodology regarding the upcoming applications; Calculating the similarity}\label{app.3}
\begin{algorithmic}

\State \textbf{2) Implementing a downstream task from the list A.}
\If{Task := \textit{Clf}}
    \SplitState {i) Implement a best baseline classifier model on the original dataset,}
    \SplitState{ii) Extract the important features affecting the final output of the model using SHAP values,}
    \SplitState{iii) Define the Privileged/Unprivileged groups based on the sensitive features in the output of the (ii) step, }
    \SplitState{iv) Calculate, TPR, FPR (or other fairness metrics) for these groups.}

    \While{($|TPR_{Privileged}-TPR_{Unprivileged}|>\delta$ or \SplitState{$|FPR_{Privileged}-FPR_{Unprivileged}|>\delta$)}}
    \SplitState {i) Map the dataset into similarity space using Algorithm \ref{app.1}, or tuning the kernels.}
    \SplitState {ii) Implement a secondary classifier on the mapped data}

\If{($|TPR_{Privileged}-TPR_{Unprivileged}|<\delta$ or \SplitState{$|FPR_{Privileged}-FPR_{Unprivileged}|<\delta$)}}
\SplitState{\textit{The dataset could be certified for the measured metrics.}}
\EndIf
\EndWhile

\If{Task:= \textit{Imp}}
\For{$i \in  N$}
\For{$j \in neighbor(i)$}
\SplitState {for every missing feature ($m_{i}$) use the  Aggregation function $(F(m_{i},m_{j}))$}
\EndFor
\EndFor
\EndIf

\If{Task:= \textit{Aug}}
\SplitState {i) Finding the number of components (c) and instances ($n_{c}$) in each one}
\SplitState{ii) Starting from nodes in the smallest component of similarity network ($c_{i}$)}
\For{all \{$n_{l},n_{k}$\} $\in n_{c}$}
\SplitState{i) generate a node assign the average of features of $n_{l},n_{k}$}
\SplitState{ii) Run the [\texttt{doubleblind}] algorithm as Vector-label propagation for labeling the synthetic nodes}
\EndFor
\EndIf
\State To certify the augmented dataset fairness, the Task:=\textit{Clf} could be implemented measuring the metrics.
\EndIf

\end{algorithmic}
\end{algorithm*}
\end{appendices}
\end{document}